\documentclass[runningheads]{llncs}

\usepackage[T1]{fontenc}

\usepackage{graphicx}
\usepackage{comment}
\usepackage{amsmath,amssymb}
\usepackage{color}
\usepackage{url}
\usepackage{hyperref}
\usepackage[dvipsnames]{xcolor}
\usepackage{fixcaption}
\usepackage{subcaption}
\usepackage{amsmath}
\usepackage{amssymb}
\usepackage{booktabs}

\usepackage[acronym,nomain,nonumberlist]{glossaries}
\newacronym{CT}{CT}{Computed Tomography}
\newacronym{mri}{MRI}{Magnetic Resonance Imaging}
\newacronym{log}{LoG}{Laplacian of Gaussian}
\newacronym{RMSE}{RMSE}{Root Mean Square Error}
\newacronym{3D}{3D}{3-dimensional}
\newacronym{2D}{2D}{2-dimensional}
\newacronym{1D}{1D}{1-dimensional}
\newacronym{LoG}{LoG}{Laplacian of Gaussian}
\newacronym{DoG}{DoG}{Difference of Gaussians}

\newif\ifreview
\reviewfalse

\ifreview
	\usepackage{lineno}

	\linenumbers
\fi

\begin{document}


\def\SubNumber{57}

\def\GCPRTrack{Main Track}

\title{Towards synthetic generation of realistic wooden logs}

\ifreview
\else

	\author{Fedor Zolotarev\inst{1}\orcidID{0009-0008-1978-5764} \and
Bo\v{r}ek Reich\inst{1, 2}\orcidID{0000-0002-5743-7671} \and
Tuomas Eerola\inst{1}\orcidID{0000-0003-1352-0999} \and
Tomi Kauppi\inst{3} \and
Pavel Zemcik\inst{1, 2}\orcidID{0000-0001-7969-5877}}
\authorrunning{F. Zolotarev et al.}
%
\institute{Lappeenranta-Lahti University of Technology LUT,
P.O. Box 20, FI-53851 Lappeenranta, Finland\\
\email{fedor.zolotarev@lut.fi}, \email{borek.reich@lut.fi},\\
\email{tuomas.eerola@lut.fi}, \email{pavel.zemcik@lut.fi} \and
Brno University of Technology BUT, Brno, Czech Republic\\
\email{reich@fit.vutbr.cz}, \email{zemcik@fit.vutbr.cz}\\
\and
Finnos Oy, Tukkikatu 5, FI-53900 Lappeenranta, Finland\\
\email{tomi.kauppi@finnos.fi}}
\fi

\maketitle              

\begin{abstract}
In this work, we propose a novel method to synthetically generate realistic 3D representations of wooden logs. Efficient sawmilling heavily relies on accurate measurement of logs and the distribution of knots inside them. \gls{CT} can be used to obtain accurate information about the knots but is often not feasible in a sawmill environment. A promising alternative is to utilize surface measurements and machine learning techniques to predict the inner structure of the logs. However, obtaining enough training data remains a challenge. We focus mainly on two aspects of log generation: the modeling of knot growth inside the tree, and the realistic synthesis of the surface including the regions, where the knots reach the surface. This results in the first log synthesis approach capable of generating both the internal knot and external surface structures of wood. We demonstrate that the proposed mathematical log model accurately fits to real data obtained from \gls{CT} scans and enables the generation of realistic logs. 

\end{abstract}

\section{Introduction}
\label{sec:intro}
Accurate measurement of logs, i.e. the main wooden axis or stem of a tree with branches and bark removed, is essential for forestry and forest industry including sawmilling. The gold standard for log measurement is X-ray \gls{CT} providing detailed information about the internal structure of the logs. Obtaining dense \gls{CT} scans of logs is typically time-consuming and expensive rendering it unsuitable for most practical applications. 
Alternatively, it is possible to use faster and cheaper sensors, such as laser range scanners, to estimate internal log characteristics~\cite{zolotarev2020modelling}. One approach would be to rely on deep learning and use a training set of dense X-ray scans. 
This, however, leads to a new problem: how to obtain large-scale training data of dense CT scans to train the data-hungry deep learning methods. A remedy for this would be pre-training on synthetic data calling for an accurate model to generate realistic logs.

We approach the synthetic log generation from the point of view of the sawmill process. Advances in timber tracing~\cite{zolotarev2019timber} have opened novel possibilities to establish feedback loops and optimize the sawmill process. The most important factor defining the quality of timber products is the locations and types of knots. Therefore, the main objective in sawmill process optimization is to select the sawing parameters that lead to a reduction of knot occurrences on the edges of the boards and this way maximize their value. One of the most commonly used log measurements in sawmills is the laser scans of the log surface. Studies on modeling the internal knot distribution based on the surface features of logs have been already carried out~\cite{zolotarev2020modelling,xiu2023modelfromsurface,duchateau2013modelling}. To further develop these models and to establish a sawing optimization based on surface measurements, we need a synthetic log model that is able to accurately model both the internal structure of knots and the log surface.

\begin{figure}[htb!]
  \centering
  
\begin{subfigure}{0.7\linewidth}
    \includegraphics[width=1\linewidth]{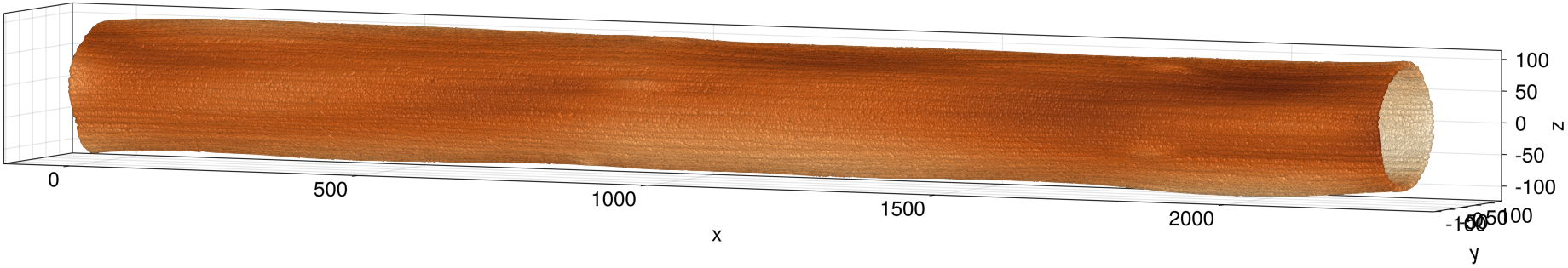}
    \caption{Generated log}
    \label{subfig:gen_log_full}
  \end{subfigure}
  \begin{subfigure}{0.7\linewidth}
    \includegraphics[width=1\linewidth]{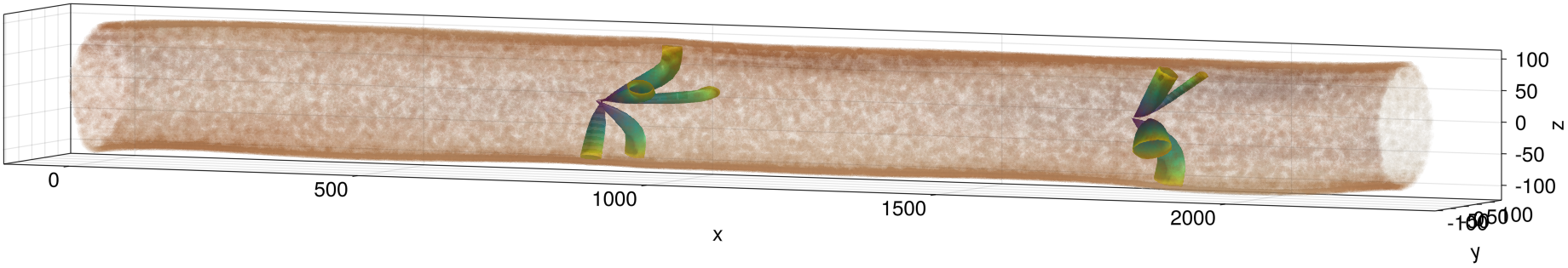}
    \caption{Generated knots}
    \label{subfig:gen_log_knots}
  \end{subfigure}
   \caption{An example of a generated log using the proposed model.}
   \label{fig:generatedLogExample}
\end{figure}

In this work, a novel log model for synthetic data generation is proposed focusing mainly on two aspects: 1) an accurate knot model that defines the growth inside the log, and 2) a realistic surface model that simultaneously characterizes the base shape of the log, knot locations on the surface, and the grain texture. This is achieved by dividing the synthetic log generation into three steps: 1) internal knot generation, 2) centerline generation, and 3) surface generation, each with a separate model. We propose using a log-centric coordinate space to separate the centerline from the surface parameters and to represent the surface via heightmap defining the distances of the surface points to the centerline. We further propose a novel geometric knot model based on the analysis of CT scans of Scots pine logs. Our log surface model comprises three components: the base shape, surface knots, and the surface grain, which control the surface details at different level scales. Finally, we derive a method to fit the model to real data to demonstrate the accuracy of the model and to obtain statistics of the model parameters for synthetic log generation.

The proposed log model enables the synthetic generation of artificial logs, allowing for the development and pretraining of various surface-based log measurement and sawing process optimization methods, such as surface knot detection, internal knot distribution modeling, virtual sawing, and sawing angle optimization methods. The main contributions of this work are as follows: 1) constructing a novel knot model using X-ray reconstructions of real logs to provide a more accurate parameterization of the knot shape, 2) a novel log surface model that simultaneously characterizes the base shape of the logs, surface knots, and the grain pattern, and 3) the first realistic \gls{3D} model for synthetic generation of both the internal knot structure and the surface of wooden logs. The source code is available at \href{https://github.com/Dysthymiac/SyntheticLogs.jl}{\color{blue}\rmfamily https://github.com/Dysthymiac/SyntheticLogs.jl}.

\section{Related work}
\label{sec:related}

\subsection{Synthetic log generation}
While a large body of work on the synthetic generation of trees exists, particularly the branching structures, from the computer graphics perspective (see e.g.~\cite{xie2015tree}), the synthetic generation of fine-scale surface and internal structures of wooden logs or stems without external branches has not been widely studied in the literature. Most existing works utilize very simplified log models to develop and test specific methods. For example, in~\cite{senchukova2023geometry}, digital phantoms of \gls{2D} slices of logs were used to evaluate parameter estimation for sparse X-ray log imaging. The digital phantoms consisted of concentric rings that coarsely modeled tree growth rings and ellipses that represented knots. In~\cite{norell2009creating}, a method to generate artificial log end images was proposed emphasizing realistic annual rings, heartwood, and knots. Both of the methods generate only \gls{2D} slices instead of the full \gls{3D} model, which further limits their usability. 

\subsection{Knot growth modeling}
An essential requirement for the realistic generation of synthetic logs is accurate geometric models that capture the log characteristics. Since the quality of the end product is largely determined by the number, locations, and types of knots, the vast majority of geometric models for wooden logs focus on modeling the knot morphology and growth inside logs. For example, in~\cite{duchateau2013modelling}, a Weibull function-based knot geometry model was proposed, complemented by additional equations for the diameter and curvature of the knot along its pith. The authors further utilized a tree growth simulator to model the knot parameters as a function of external branch and tree characteristics. In~\cite{guo2022modeling}, the knots were divided into two categories based on their shape (curved and linear knots), and different models were used for the categories.
Authors of~\cite{nordmark2003models} defined knot's radius, longitudinal and tangential positions as functions of the distance to the pith. It is also worth noting that they have used similar log-centric coordinates and also modeled the surface of the log as well.

\subsection{Log surface modeling}
Several methods for tree stem shape have been proposed for biomass and size distribution estimates based on terrestrial laser scanning (TLS). The simplest geometric primitive to model the log surface is the circular cylinder but more complex primitives have also been suggested. For example, in~\cite{aakerblom2015analysis}, the elliptical cylinder, circular cone, polygonal cylinder, and polyhedral cylinder surfaces were compared on TLS data. Using simple geometric primitives, however, fails to capture the natural variation of the shape as well as the fine-grained surface details, such as knots. This renders the shape models from TLS literature unsuitable for sawmill applications, where detection and analysis of surface knots and other defects is essential.

Wood and tree bark texture synthesis, including \gls{3D} bark generation~\cite{venkataramanan2022data,lefebvre2002bark,liu2016wood}, has been extensively covered in computer graphics literature. Such methods, however, typically aim for an aesthetically pleasing result instead of accurate surface models. 
For example, in~\cite{xie2015tree}, geometric textures were transferred from an existing database to the generated tree surface. One interesting work was presented in~\cite{liu2015procedural}, where various elements of wood, such as longitudinal and ray fibers, were analyzed, and based on this a randomized approach for wood texture synthesis was proposed. The method utilizes sparse convolution noise~\cite{lagae2009procedural} to produce a similar texture to the wood surface.

All of the existing log and tree stem models as well as synthetic tree generation methods focus on one or few aspects of the problem, such as the internal structure or the surface. To the best of our knowledge, no synthetic log model that combines both the geometric knot model and surface model exists.

\section{Synthetic log model}
\label{sec:proposed}

\subsection{Model components}
\label{sec:components}

The goal is to describe the surface of a debarked log as well as the internal distribution of knots. The proposed log model consists of several components each describing separate parts that are combined into a final \gls{3D} log shape. Using a similar approach to the methods from~\cite{zolotarev2019timber,zolotarev2020modelling,batrakhanov2021virtual}, the model is defined with the log-centric coordinate space, allowing to separate the centerline of the log from the rest of the surface parameters. Thus, in order to create a synthetic log, it is necessary to generate a random curve that will serve as the centerline and a corresponding heightmap that will completely describe the surface of the log. However, since many of applications~\cite{zolotarev2019timber,zolotarev2020modelling,batrakhanov2021virtual} that made the use of \gls{3D} log data necessary are working with surface defects, mainly knots, it is necessary to model them as well. Therefore, the whole synthetic log model can be divided into three main components: 1) knots, 2) centerline, and 3) surface.

\subsection{Knot model}
\label{sec:knotModel}

\begin{figure}[htb!]
  \centering
  \begin{subfigure}{0.24\linewidth}
    \includegraphics[width=1\linewidth]{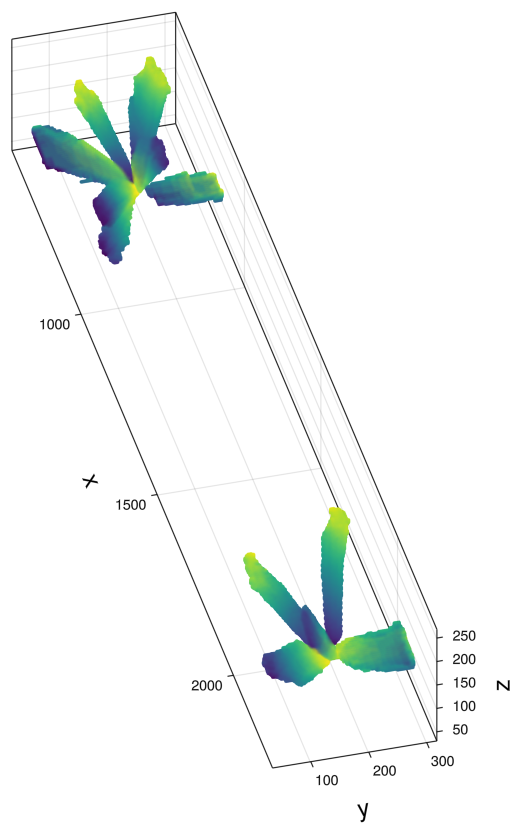}
    \caption{}
    \label{subfig:cartesian}
  \end{subfigure}
  \begin{subfigure}{0.35\linewidth}
    \includegraphics[width=1\linewidth]{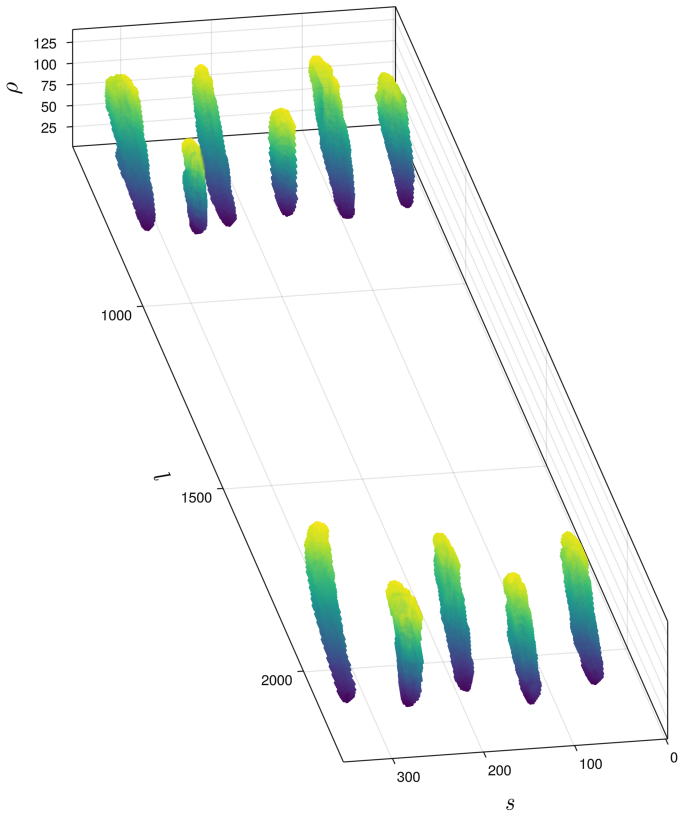}
    \caption{}
    \label{subfig:logcentric}
  \end{subfigure}
   \caption{Comparison of knot points in (\subref{subfig:cartesian})~Cartesian and (\subref{subfig:logcentric})~Modified log-centric coordinates. Axis $s$ is in degrees. }
   \label{fig:cartesianVsLogcentric}
\end{figure}

The proposed knot model builds on the model proposed in~\cite{duchateau2013modelling}. Similarly to the earlier model, the knot vertical position and knot radius are described with the same equation. The models for individual knots are described in the modified log-centric coordinates from~\cite{zolotarev2020modelling}. Log-centric coordinates consist of an angle around the centerline $\theta$,  longitudinal position relative to the log centerline $l$, and distance from the centerline $\rho$. In order to preserve the shape of the knot, instead of angle $\theta$, an arc length coordinate is used:
\begin{equation}
s=(\theta - \bar{\theta}) \rho,    
\end{equation}
where $\theta$ is the angle around the centerline in radians, $\bar{\theta}$ is the mean angle of all knot points and $\rho$ is the distance from the centerline. An example comparison of knots in Cartesian and modified log-centric coordinates is presented in Fig.~\ref{fig:cartesianVsLogcentric}. 

Knot vertical position from~\cite{duchateau2013modelling} is described as
\begin{equation}
\label{eq:alpha_l}
\alpha_l = \delta_l - L_l \rho_\mathrm{max},
\end{equation}
\begin{equation}
\label{eq:knotModel_l}
    K_l(\rho) = \alpha_l(1-e^{(-E_l\frac{\rho}{\rho_\mathrm{max}-\rho})})+L_l \rho,
\end{equation}
where $\delta_l$ is the difference in $l$ coordinates between the position where the knot reaches the surface and its origin, $\rho_\mathrm{max}$ corresponds to the radius $\rho$ at which the knot reaches the surface, $L_l$ and $E_l$ are linear and exponential coefficients respectively. The idea is to divide the knot lifetime into periods of exponential growth in the beginning and linear growth closer to the surface. For greater flexibility, the term $\rho_\mathrm{max}-\rho$ is replaced with $\max{(0, \rho_\mathrm{max}-\rho)}$ to allow the function to be defined after it reaches $\rho_\mathrm{max}$.

The same function is used to describe the change of the knot radius as it grows:
\begin{equation}
\label{eq:alpha_r}
\alpha_r = r_\mathrm{max} - L_r c_\mathrm{max},
\end{equation}
\begin{equation}
\label{eq:knotModel_r}
    K_r(c) = \alpha_r(1-e^{(-E_r\frac{c}{c_\mathrm{max}-\rho})})+L_r \rho,
\end{equation}
where $r_\mathrm{max}$ is the radius of the knot at the surface, $c_\mathrm{max}$ corresponds to the length of the knot at which it reaches the surface, $L_r$ and $E_r$ are linear and exponential coefficients respectively. Comapred to~\cite{duchateau2013modelling}, the radius is modeled as $K_r(c)$, where $c$ is the length of the curve described by $K_l(\rho)$, instead of being dependent on the distance to the centerline $\rho$. Consequently, the radius corresponds to a circle with a center on the knot center curve and is perpendicular to the curve. With this change, the knot radius depends on the actual growth trajectory. Another limitation of the model from~\cite{duchateau2013modelling} is that the knots are only modeled using their \gls{2D} profiles. For the purpose of the generation of a full model, it is necessary to account for their \gls{3D} shape. Following the ovality assumption from~\cite{lemiux1997shape,duchateau2013modelling}, the radius of a knot is defined as a ratio of a vertical to the horizontal semi-axis of the knot, $\gamma$. 

The model has been re-parameterized to replace the parameters $L$ and $E$ with new parameters with clearly defined boundaries. One such parametrization is to instead define the incline angles at the origin $K_l(0)$ and at the endpoint $K_l(\rho_\mathrm{max})$. 
A couple of conditions exist that need to be satisfied:
\begin{enumerate}
    \item The multiplier $\alpha$ must be non-negative. 
    \item The function must be concave, i.e. the growth rate can only decrease after the explosive growth in the beginning.
\end{enumerate}
This allows to easily specify sensible bounds for those parameters and reparametrize those angles as ratios $\phi_0, \phi_1 \in [0, 1]$ between minimum and maximum possible incline. The same reparametrization is applied to the knot radius $K_r$ model as well. The full derivation of the final knot equation can be found in the supplementary materials.

To summarize, a single knot can be described with its origin position $(s_0, l_0)$, the ovality multiplier $\gamma$, three parameters for $K_l(\rho)$: $\rho_\mathrm{max}, \phi_0, \phi_1$ and three parameters for $K_r(c)$: $r_\mathrm{max}, \psi_0, \psi_1$ , totaling 9 parameters for a single knot. Values of $\delta_l$ and $c_\mathrm{max}$ are computed using all other parameters and log surface.

\subsection{Centerline model}

Despite being a generally arbitrary curve, the centerline of a log is usually close to a straight line with somewhat rare exceptions when the tree is curved. Since self-intersection is impossible as well as turning the direction of growth for more than 90 degrees, it is possible to model the centerline as a function of $x$ instead of a more general parametric curve. That means that the centerline can be modeled by defining two functions $y(x)$ and $z(x)$. A simple option for parametrizing those functions is to approximate them using Chebyshev polynomials as basis. Since the centerline does not usually contain a lot of high-frequency variations, it is enough to use a relatively small number of coefficients from Chebyshev space to approximate the general shape. Additionally, by assuming that the coordinates of the centerline are centered such that $\bar{y}=\bar{z}=0$, it is also possible to omit the first coefficient since it is just a constant offset. This allows to model centerline with $2n$ coefficients: $n$ for $y(x)$ and $n$ for $z(x)$.
 
\subsection{Log surface model}

\subsubsection{Log thickness.}

\begin{figure}[htb!]
  \centering
  \includegraphics[width=0.7\linewidth]{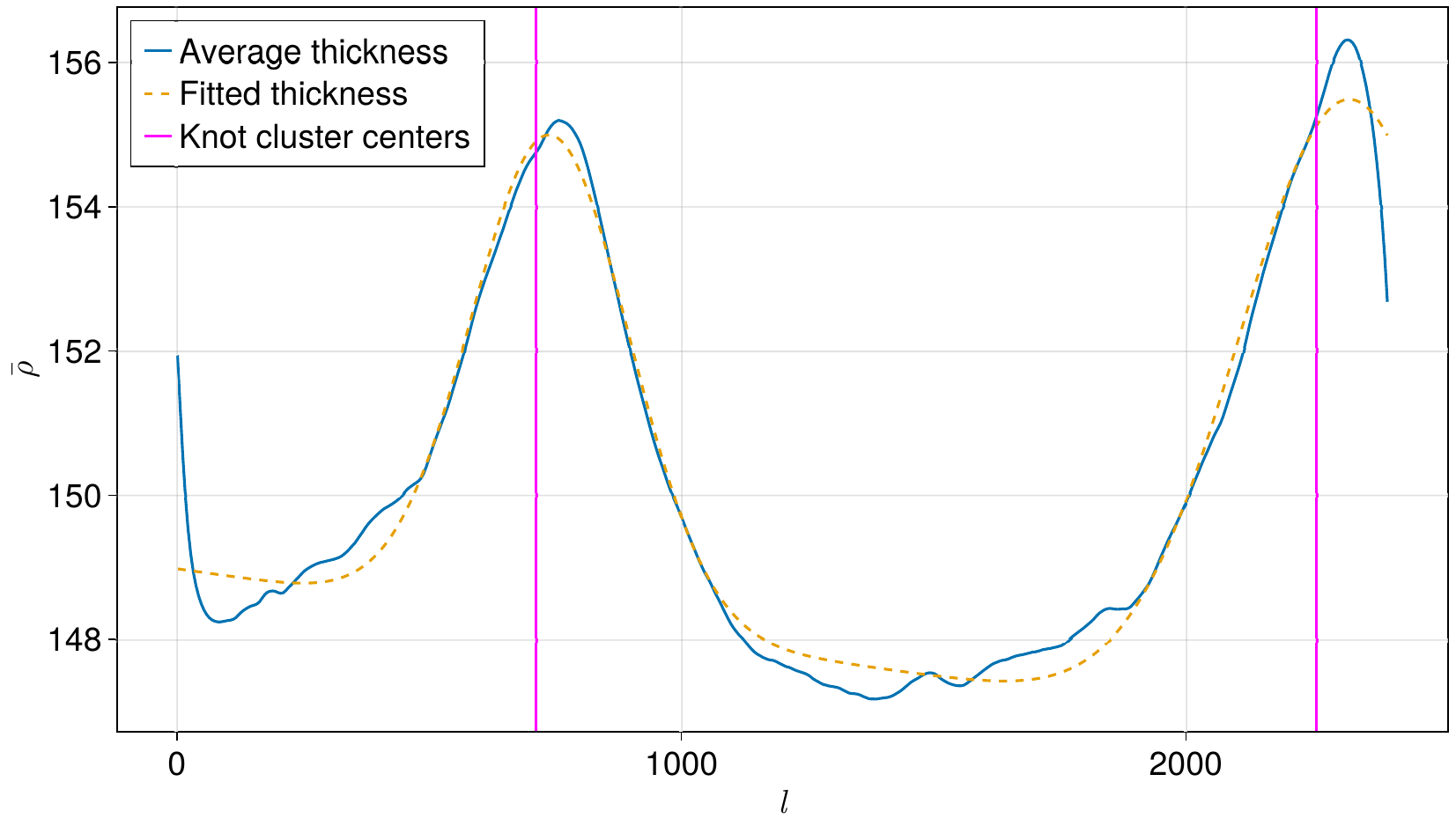}
   \caption{Plot of log thickness along the length of a log. Vertical lines correspond to the centers of knot clusters. Fitted function is plotted with dashed line.}
   \label{fig:thickness}
\end{figure}

Log thickness at height $l$ can be defined as an average radius for all angles at that height. Generally, logs are thicker at the bottom and taper to the top - a natural consequence of the growing process. Additionally, the logs are thicker in places where the knot clusters grow. An example plot of log thickness is presented on Fig.~\ref{fig:thickness}. It is possible to explicitly model average log thickness by fitting a line and a scaled Gaussian to each knot cluster location. The thickness model becomes

\begin{equation}
    \mathrm{thickness}(l) = \mathrm{a} \cdot l + \mathrm{b} + \sum_{c_i} \alpha_i \exp\left(-\frac{x-(c_i+\beta_i)^2}{2\gamma_i^2}\right),
\end{equation}
where $\mathrm{a}$ and $\mathrm{b}$ are growth rate (slope) and starting thickness (intercept), $c_i$ are cluster positions on axis $l$, $\alpha_i, \beta_i, \gamma_i$ are parameters for Gaussians corresponding to each knot cluster.

\subsubsection{Surface curvature.}

The base shape of the log, i.e. the low-frequency features of the surface heightmap, are modeled in a similar manner to the centerline. There are a couple of properties of the heightmap that are worth considering: 
    1) the heightmap is cyclic in the $\theta$ dimension due to it corresponding to an angle around the log, and
    2) barring grain, noise, and various defects, it is assumed that the general log shape is smooth.

By considering the log shape as a result of a closed smooth curve that changes in time as the tree is growing, the shape is modeled in two steps. First, each row of a heightmap is approximated using a Fourier basis, to ensure that it is periodic around the log's circumference. Since high-frequency features are not the target, only $n$ first coefficients are needed. The second step is to describe the change of those curves as the log grows, i.e. describe the change along the $l$ dimension. This is done by fitting the curves to each of the $n$ coefficients as they change along the $l$ dimension. Those curves, in turn, can be approximated using $m$ coefficients Chebyshev polynomials. This results in $m \times n$ coefficients needed to describe the surface. 

\subsubsection{Surface knots.}

\begin{figure}[htb!]
  \centering
  \begin{subfigure}{0.3\linewidth}
    \includegraphics[width=1\linewidth]{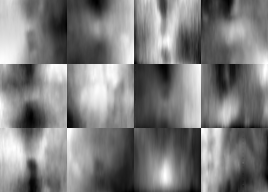}
    \caption{Surface knots}
    \label{subfig:surfaceKnots}
  \end{subfigure}
  \begin{subfigure}{0.3\linewidth}
    \includegraphics[width=1\linewidth]{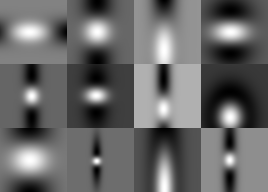}
    \caption{Fitted knots}
    \label{subfig:surfaceDoGs}
  \end{subfigure}
  \begin{subfigure}{0.3\linewidth}
    \includegraphics[width=1\linewidth]{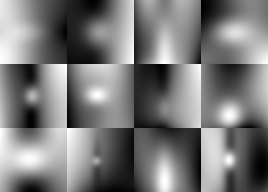}
    \caption{Reconstruction}
    \label{subfig:surfaceRecs}
  \end{subfigure}

   \caption{Surface knots (\subref{subfig:surfaceKnots}) as seen on the heightmap, (\subref{subfig:surfaceDoGs}) fitted models of surface knots and (\subref{subfig:surfaceRecs}) fitted knots on the reconstruction.}
   \label{fig:surfaceKnots}
\end{figure}

One of the most important features of the logs is the knots. It was stated in~\cite{zolotarev2020modelling} that surface knots can be recognized on the heightmap as bumps, i.e. small elevated regions and they were segmented by using \gls{LoG}. The regions of the heightmap, where the knots grow to the surface are presented in Fig.~\ref{subfig:surfaceKnots}. Apart from sometimes being visible as bumps on the surface, the knots are also accompanied by depressions around them, mostly on the vertical $l$ axis. This further confirms \gls{LoG} as a valid shape descriptor for them. However, we use a slightly modified \gls{DoG} due to it being a most common approximation to \gls{LoG} and providing more control in its construction. The 1D variant has the following form:
\begin{equation}
    K(x; \sigma_1, \sigma2, m) = m e^{-\frac{x^2}{2\sigma_1^2}} - e^{-\frac{x^2}{2\sigma_2^2}}, 
\end{equation}
where $\sigma_1, \sigma_2$ are standard deviations of Gaussians and $\sigma_1^2 > \sigma_2^2$, $m$ is an additional multiplier that controls the proportion of the elevation versus depression of the knot. It is possible to find the intersection of the fitted knot model with the surface, in that case, the size of the intersection is used to determine the radius of the surface knot bump. Since it is assumed that $\sigma_2 < \sigma_1$, the second standard deviation can be represented with a multiplier $\alpha$:
\begin{equation}
    \sigma_2 = \alpha \sigma_1,
\end{equation}
where $\alpha \in (0, 1)$.
The radius of the bump corresponds to the intersection with the $x$-axis. $\sigma_1$ can be defined in terms of the radius $r$:
\begin{equation}
    \sigma_1 = \sqrt{\frac{r^2 (\frac{1}{\alpha^2} - 1)}{2 \ln{m}}}.
\end{equation}

With these reparametrizations, the knot function $K$ can be defined by the radius $r$, the multiplier of the standard deviation $\alpha$, and the multiplier of a Gaussian $m$. The radius is defined by the internal knot model, which leaves only $\alpha$ and $m$ to be defined for the generation of a surface knot. The surface knots are defined as \gls{2D} images, which means that the multiplier $\alpha$ is a tuple of multipliers for the axes $l$ and $\theta$ and results in the surface knot being defined by only three scalars. An example of the fitted models is presented in Fig.~\ref{subfig:surfaceDoGs}
.

\subsubsection{High frequency features.}

Sparse Gabor convolution noise~\cite{lagae2009procedural} has been used to generate the high-frequency details of the log surface. Similarly, this approach has been used in~\cite{liu2015procedural} to model anatomical properties of the wood, such as longitudinal and ray fibers. In the proposed model, high-frequency features are modeled by convolving the Gaussian noise with the Gabor filter. Since the majority of cells grow in the longitudinal direction~\cite{hoadley2000understanding}, the orientation of the Gabor filter was chosen to reflect that property. To increase the amount of details on the reconstruction, the noise was generated in several octaves and summed up. 

\section{Experiments}
\label{sec:experiments}
To evaluate the proposed log model, data from X-ray reconstructions of real logs were used. The purpose of the experiments was to test how accurately the model can fit real data, and how well it can reconstruct the logs. Finally, a qualitative analysis of the synthesis method's capability to generate novel logs was carried out.

\subsection{Data}

The data consists of X-ray \gls{CT} reconstructions of 8  
Scots pine (\textit{Pinus sylvestris}) logs with manually annotated knots. 
The reconstructions were also scaled to correspond to the real size of the log in millimeters. X-ray reconstructions were thresholded to remove the background and log surface was used to create surface point clouds.

Since X-ray density was used to threshold the log, some high-density bark segments might still be present on the point clouds.

\subsection{Knot model}
\subsubsection{Model fitting.}

The model described in Sec.~\ref{sec:knotModel} was fitted to the points corresponding to the knots surface. Initial parameters are found by using least squares to individually fit all knot parameters, i.e. estimating all incline angles and points. Please refer to the supplementary materials for more details about the initial fitting. After all initial parameters are found, they are further refined using Levenberg-Marquardt method~\cite{levenberg1944method}.

\begin{figure}[htb!]
  \centering
  \begin{subfigure}{0.4\linewidth}
    \includegraphics[width=1\linewidth]{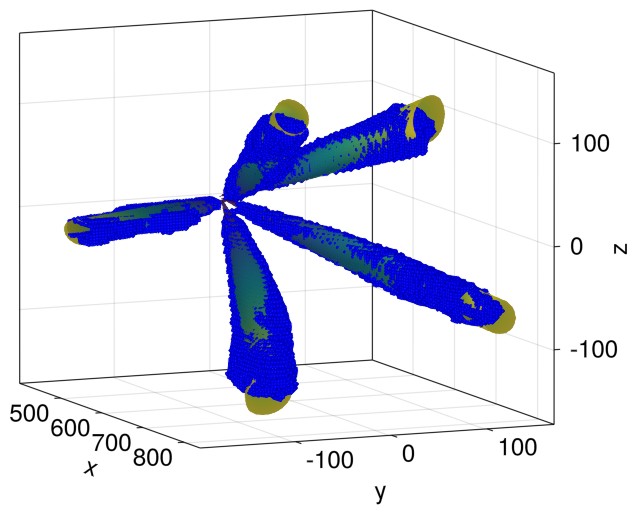}
    \caption{}
    \label{subfig:fittedCluster}
  \end{subfigure}
  \begin{subfigure}{0.4\linewidth}
    \includegraphics[width=1\linewidth]{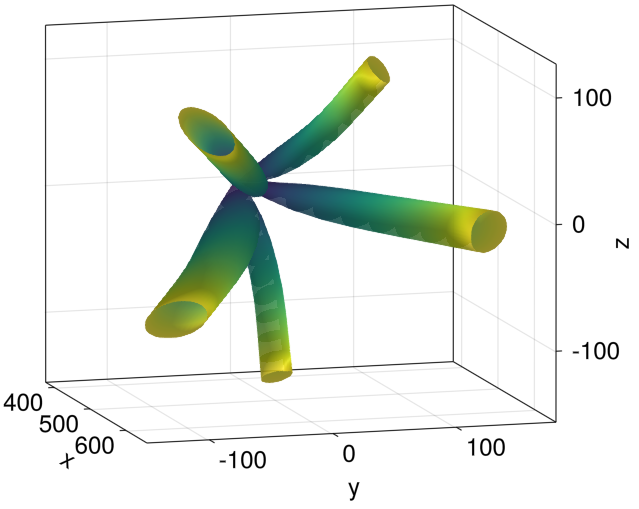}
    \caption{}
    \label{subfig:generatedCluster}
  \end{subfigure}

   \caption{Knot model: (a) fitted models for a cluster of knots; (b) generated knot cluster.}
   \label{fig:knot_model_exp}
\end{figure}

The values of \glspl{RMSE} for all fitted knot models are presented in Table~\ref{tab:rmse}. A plot of fitted knot models along with knot points is presented in Fig.~\ref{subfig:fittedCluster}. 

\begin{table}[t]
\begin{center}
\caption{\glspl{RMSE} of knot model fitted to the knot data. The measurements are in millimeters.}
\label{tab:rmse}
\begin{tabular}{lp{0.5cm}p{1cm}p{1cm}p{1cm}p{1cm}p{1cm}p{1cm}p{1cm}p{1cm}l}
  \toprule
\textbf{Log} & & Log 1 & Log 2 & Log 3 & Log 4 & Log 5 & Log 6 & Log 7 & Log 8 & All logs \\
\midrule   
\textbf{Knots} & & 12 & 9 & 12 & 15 & 18 & 13 & 10 & 13 & 102 \\
\textbf{RMSE} & \texttt{$\mu$} & 4.78  & 3.99  & 5.72  & 5.71  & 5.84  & 3.99  & 3.03  & 3.81  & 4.75 \\
 & \texttt{$\sigma$} & 0.79 & 1.36 & 1.31 & 1.57 & 1.51 & 0.78 & 0.29 & 0.97 & 1.52
\\\bottomrule
\end{tabular}
\end{center}
\end{table}

\subsubsection{Generation.}
In order to model the distribution of knot parameters, 
compound Gaussian distribution has been chosen due to its relative simplicity and great versatility. The final knot parameters $X$ are distributed as
\begin{equation}
    X \sim \mathcal{N}(\mu, (\sigma \sigma^\intercal) \odot \mathrm{P}),
\end{equation}
where $X$ is the random vector corresponding to knot parameters, $\mu$ is the mean vector, $\sigma$ is the standard deviation vector, $\odot$ is the Hadamard product, and $\mathrm{P}$ is the correlation matrix of knot parameters. In turn, $\mu \sim \mathcal{N}(\mu_1, \sigma_1)$ and $\sigma \sim \mathcal{N}(\mu_2, \sigma_2)$, and so on for $\mu_1, \sigma_1, \mu_2, \sigma_2$.

By modeling distributions within a log, within a knot cluster, and a knot as separate random variables with clear dependencies, i.e. knots within a cluster must share similar characteristics but two knots taken from different clusters can be different. An example of a generated knot cluster is presented in Fig.~\ref{subfig:generatedCluster}.

\subsection{Log surface model}

\subsubsection{Reconstruction.}

\begin{figure}[htb!]
  \centering
  \begin{subfigure}{0.23\linewidth}
    \includegraphics[width=1\linewidth]{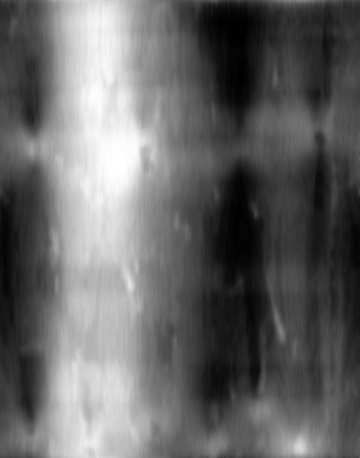}
    \label{subfig:heightmap}
  \end{subfigure}
  \begin{subfigure}{0.23\linewidth}
    \includegraphics[width=1\linewidth]{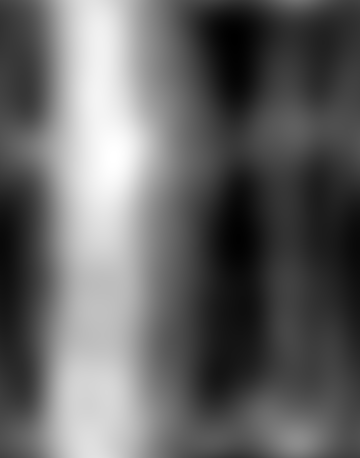}
    \label{subfig:lf_heightmap}
  \end{subfigure}
  \begin{subfigure}{0.23\linewidth}
    \includegraphics[width=1\linewidth]{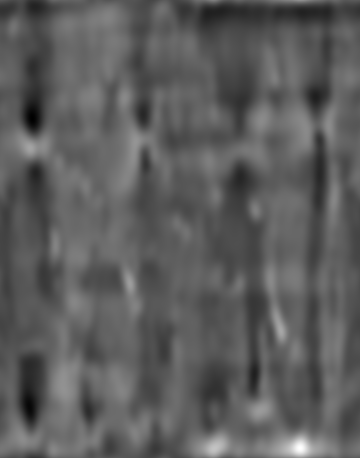}
    \label{subfig:mf_heightmap}
  \end{subfigure}
  \begin{subfigure}{0.23\linewidth}
    \includegraphics[width=1\linewidth]{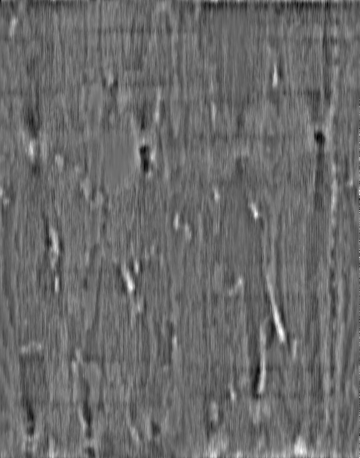}
    \label{subfig:hf_heightmap}
  \end{subfigure}
  \\ 
  \begin{subfigure}{0.23\linewidth}
    \includegraphics[width=1\linewidth]{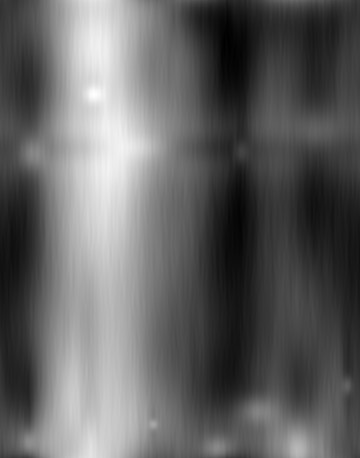}
    \caption{}
    \label{subfig:rec}
  \end{subfigure}
  \begin{subfigure}{0.23\linewidth}
    \includegraphics[width=1\linewidth]{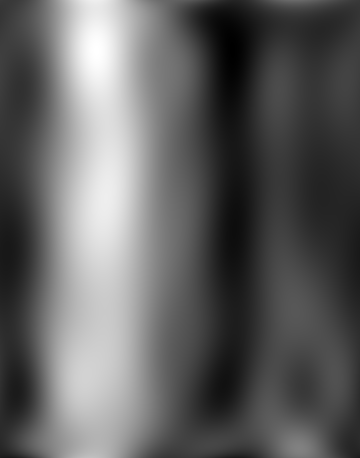}
    \caption{}
    \label{subfig:lf_rec}
  \end{subfigure}
  \begin{subfigure}{0.23\linewidth}
    \includegraphics[width=1\linewidth]{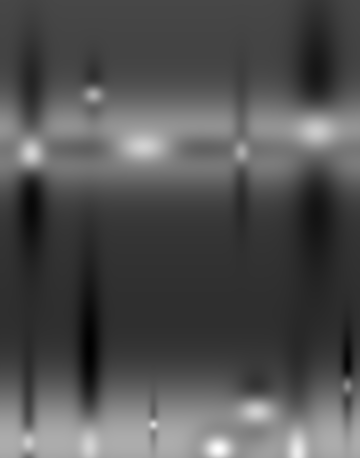}
    \caption{}
    \label{subfig:mf_rec}
  \end{subfigure}
  \begin{subfigure}{0.23\linewidth}
    \includegraphics[width=1\linewidth]{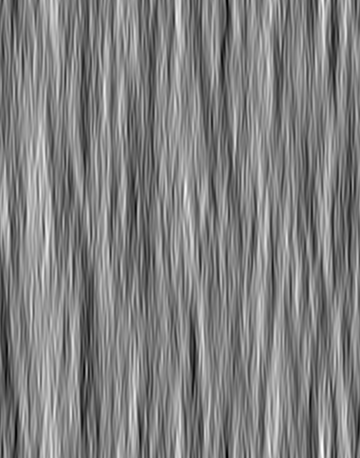}
    \caption{}
    \label{subfig:hf_rec}
  \end{subfigure}
  
   \caption{Comparison of different bands of measured heightmap frequencies (top row) with components of the reconstructed heighmap (bottom row): (\subref{subfig:rec})~original heightmap and full reconstruction; (\subref{subfig:lf_rec})~low-frequency details of the heightmap and reconstruction;  (\subref{subfig:mf_rec})~medium frequency details; (\subref{subfig:hf_rec})~high-frequency details.}
   \label{fig:reconstructionDetails}
\end{figure}

A comparison of the heightmap and its reconstruction is presented in Fig.~\ref{subfig:rec}. Parameter estimation begins with fitting the base shape parameters. First, each row of the heightmap $\rho(\theta)$ is approximated with 10 first Fourier coefficients, resulting in $n \times 10$ coefficients, which are treated as 10 curves $c(l)$ and in turn approximated with 10 coefficients for Chebyshev polynomials. This results in 100 coefficients total, that can be used to describe the base surface of a log. An example of such reconstruction is presented on Fig.~\ref{subfig:lf_rec}. Parameters for the Gabor filter for the high-frequency details were fixed for all available logs and were chosen empirically to match the high-frequency details of real heightmaps. An example comparison is presented on Fig.~\ref{subfig:hf_rec}. And finally, the parameters for the surface knots were found by minimizing the \gls{RMSE} between the knot on the heightmap and the knot on the reconstruction. Reconstructed knots are presented in Fig.~\ref{subfig:mf_rec}. It is worth noting that the goal of the proposed model is to describe a debarked wooden log, and it is possible that some surface details correspond to the bark and were not accounted for by the proposed model.  

\subsubsection{Generation.}

\begin{figure}[htb!]
  \centering
  \begin{minipage}[c]{0.46\textwidth}
  \begin{subfigure}{\linewidth}
    \includegraphics[width=1\linewidth]{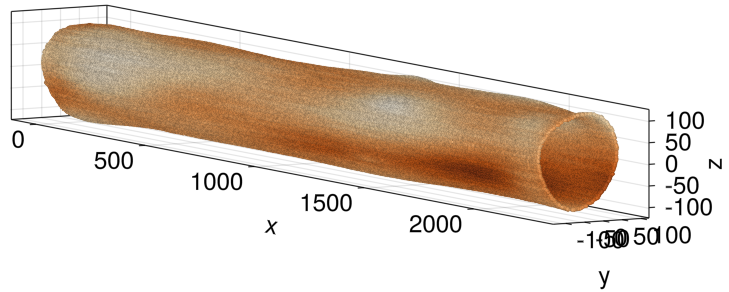}
    \caption{Generated log}
    \label{subfig:gen_log}
  \end{subfigure}
  \begin{subfigure}{\linewidth}
    \includegraphics[width=1\linewidth]{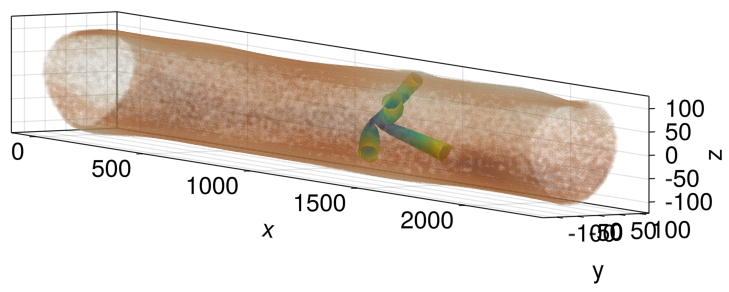}
    \caption{Generated knots}
    \label{subfig:gen_knots}
  \end{subfigure}
  \end{minipage}
  \begin{minipage}[c]{0.35\textwidth}
  \begin{subfigure}{\linewidth}
    \includegraphics[width=1\linewidth]{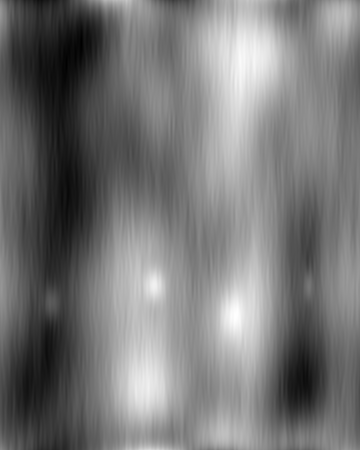}
    \caption{Generated heightmap}
    \label{subfig:gen_map}
  \end{subfigure}
  \end{minipage}
   \caption{Results of generation: (\subref{subfig:gen_log})~A full log, (\subref{subfig:gen_knots})~Generated knots, (\subref{subfig:gen_map})~A corresponding heightmap.}
   \label{fig:generatedLog}
\end{figure}

A normal distribution is fitted to surface heightmap coefficients and used to create a random surface base. Due to a limited amount of data available, a simplifying assumption that the coefficients are independently distributed is made. A multivariate normal distribution is fitted to the surface knot parameters. Since the location and size of the surface knots depend on the internal knots, they need to be generated first and the locations of their intersection with the base surface are used to initialize the surface knots with randomized parameters. Similarly, the centerline is randomized by fitting multivariate normal distributions to the available centerline coefficients. An example of a randomly generated surface is presented in Fig.~\ref{subfig:gen_map} and the corresponding \gls{3D} log is presented in Fig.~\ref{subfig:gen_log}.

\section{Conclusion}
\label{sec:conclusion}
In this paper, we describe a novel method for generating synthetic 3D wooden logs to develop and train methods for quality control and process optimization in sawmills. The proposed method is the first wooden log synthesis approach capable of simultaneously generate both the internal knot and surface structures. The method comprises separate models for the knots, centerline, and surface. This is achieved by utilizing a log-centric coordinate system that allows us to separate the centerline from the surface parameters. The proposed knot growth model consists of 9 parameters, providing an accurate fit to real knots and a realistic synthesis of artificial knots simultaneously. The log surface model utilizes separate submodels to characterize the base shape, surface knots, and other high-frequency features of the log surface. This enables the generation of realistic log surfaces with bumps corresponding to knots, similar to those on real logs. The proposed method facilitates the generation of a large number of artificial logs, allowing for the development and pretraining of various surface-based log measurement and sawing process optimization methods, such as surface knot detection, internal knot distribution modeling, virtual sawing, and sawing angle optimization methods.

\section*{Acknowledgements}
The research was supported by the Finnish Research Impact Foundation (project number 241).

%
%
%
%
\bibliographystyle{splncs04}
\bibliography{057-main.bib}

\end{document}